\documentclass[twoside]{article}

\usepackage{aistats2019}
\usepackage[round,authoryear]{natbib}

\usepackage[utf8]{inputenc} 
\usepackage[T1]{fontenc}    
\usepackage{lmodern}
\usepackage{hyperref}       
\usepackage{url}            
\usepackage{booktabs}       
\usepackage{amsfonts}       
\usepackage{nicefrac}       
\usepackage{microtype}      
\usepackage{multirow}
\usepackage{bm}

\usepackage{url}
\usepackage{graphicx} 
\usepackage{subcaption} 

\usepackage{algorithm}
\usepackage{algorithmic}

\usepackage{multirow}
\usepackage{amsmath}
\usepackage{amssymb}
\usepackage{amsthm}
\usepackage{bbm}
\usepackage{xspace}
\usepackage{enumitem}
\usepackage{color}
\usepackage{placeins} 

\usepackage{dsfont}

\usepackage{comment}
\usepackage{color}
\usepackage{bm}
\usepackage{amsthm,amsmath,amssymb}
\usepackage{array}
\usepackage{wrapfig}
\usepackage{lipsum}
\usepackage{enumitem}
\usepackage{relsize}
\setenumerate{leftmargin=*}
\setenumerate[1]{label=(\arabic*),align=left}
\setenumerate[2]{label=(\alph*)}

\newtheorem{theorem}{Theorem}

\newtheorem{definition}{Definition}

\newtheorem{remark}{Remark}

\makeatletter
\DeclareRobustCommand\onedot{\futurelet\@let@token\@onedot}
\def\@onedot{\ifx\@let@token.\else.\null\fi\xspace}
\makeatother

\def\Wlg{{W.l.o.g}\onedot}
\def\eg{{e.g}\onedot} 
\def\ie{{i.e}\onedot} 
 
\def\etc{{etc}\onedot} 
\def\wrt{w.r.t\onedot} 
\def\etc{etc\onedot}
\def\etal{{et al}\onedot}

\newcommand{\R}{\mathbb{R}}
\newcommand{\X}{\mathcal{X}}

\DeclareMathOperator*{\argmin}{\operatorname{argmin}}

\DeclareMathOperator{\dist}{dist}

\def\calC{{\cal C}}
\def\calF{{\cal F}}
\def\calD{{\cal D}}
\def\calX{{\cal X}}

\def\calT{{\cal T}}

\newcommand{\sgn}{\mathop{\mathrm{sign}}}

\numberwithin{equation}{section}
\numberwithin{theorem}{section}

\newcommand{\myparagraph}[1]{\smallskip\noindent\textbf{#1}}

\DeclareMathOperator{\dgm}{dgm}

\newcommand{\superemph}[1]{\textbf{#1}}

\newcommand{\LT}{L_{\calT}}

\begin{document}

%

%

\twocolumn[

\aistatstitle{A Topological Regularizer for Classifiers via Persistent Homology}

\aistatsauthor{ Chao Chen \And Xiuyan Ni \And Qinxun Bai \And Yusu Wang }

\aistatsaddress{ Stony Brook University \And  City University of New York \And Hikvision Research America \And Ohio State University} ]

%
%


\begin{abstract}
    Regularization plays a crucial role in supervised learning. 
Most existing methods enforce a global regularization in a structure agnostic manner.
In this paper, we initiate a new direction and propose to enforce the structural simplicity of the classification boundary by regularizing over its \emph{topological complexity}. 
In particular, our measurement of topological complexity incorporates the \emph{importance} of topological features (e.g., connected components, handles, and so on) in a meaningful manner, and provides a direct control over spurious topological structures. 
We incorporate the new measurement as a topological penalty in training classifiers. We also propose an efficient algorithm to compute the gradient of such penalty.
Our method provides a novel way to topologically simplify the global structure of the model, without having to sacrifice too much of the flexibility of the model. 
We demonstrate the effectiveness of our new topological regularizer on a range of synthetic and real-world datasets.
\end{abstract}

\section{Introduction}
\label{sec:intro}
\begin{figure*}[hbtp]
\vspace{-.1in}
  \centering
  \begin{tabular}{cccc}
  \includegraphics[height=.18\textwidth,angle=90]{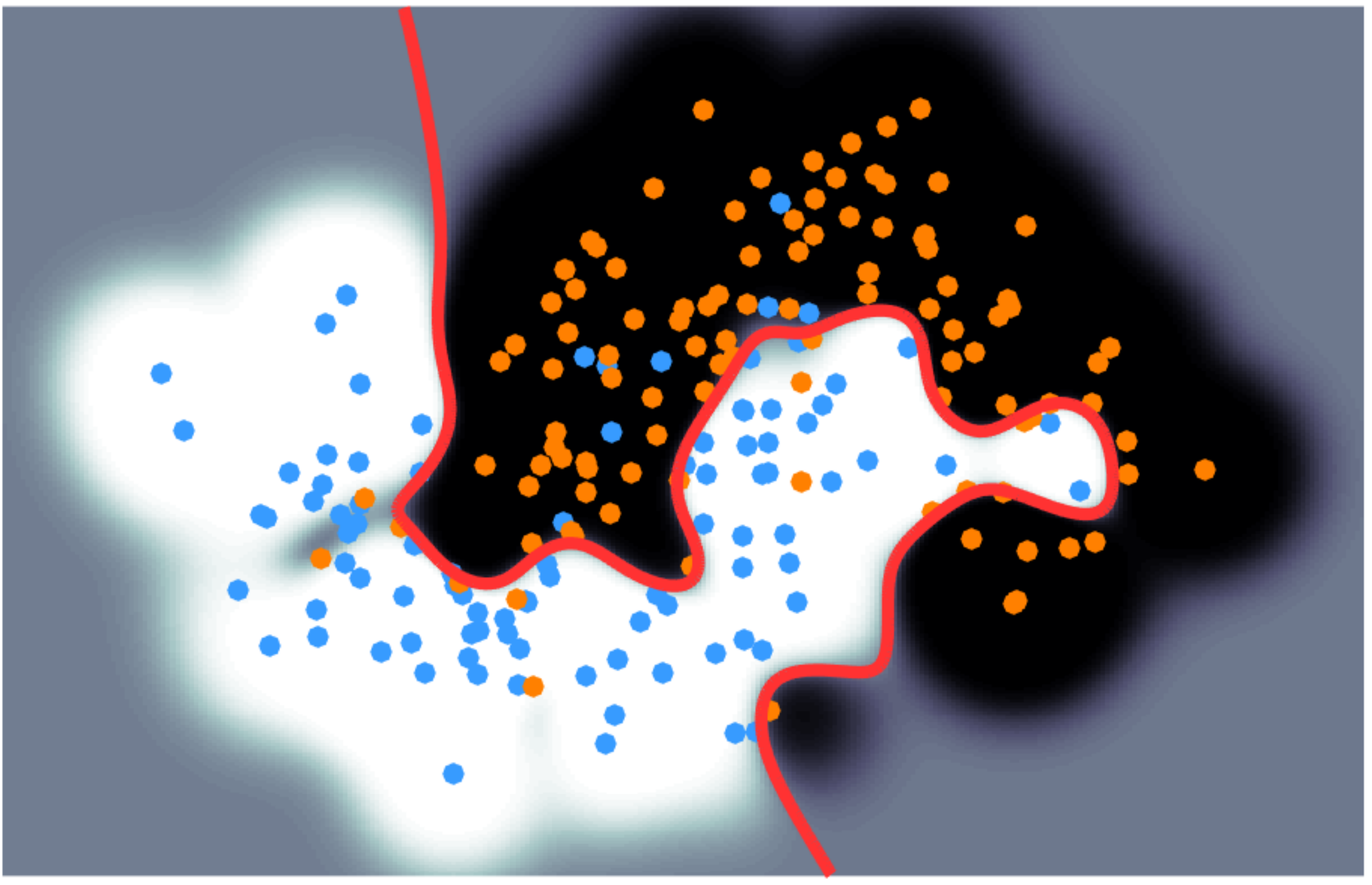} &
  \includegraphics[height=.18\textwidth,angle=90]{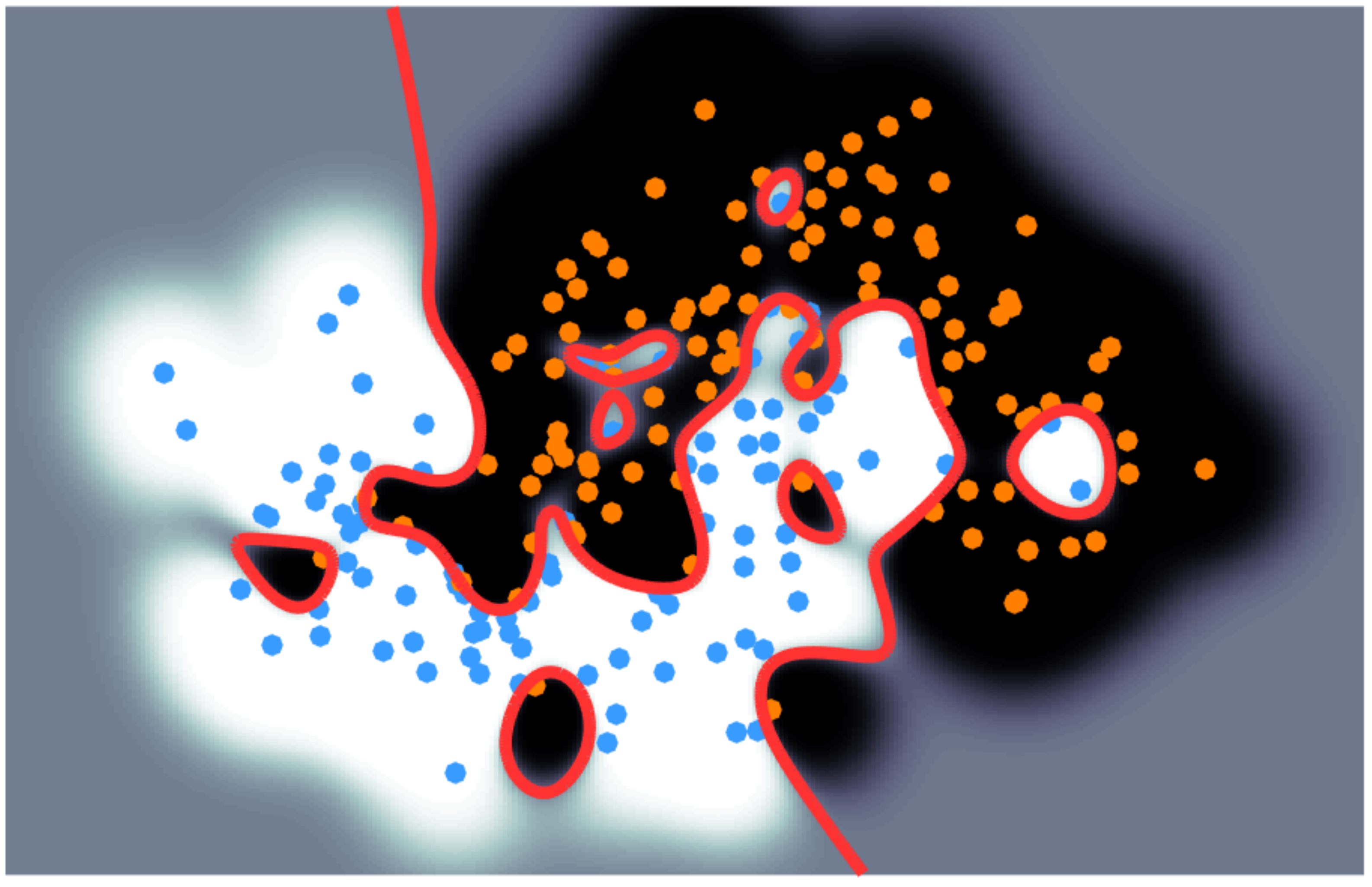} &
  \includegraphics[height=.18\textwidth,angle=90]{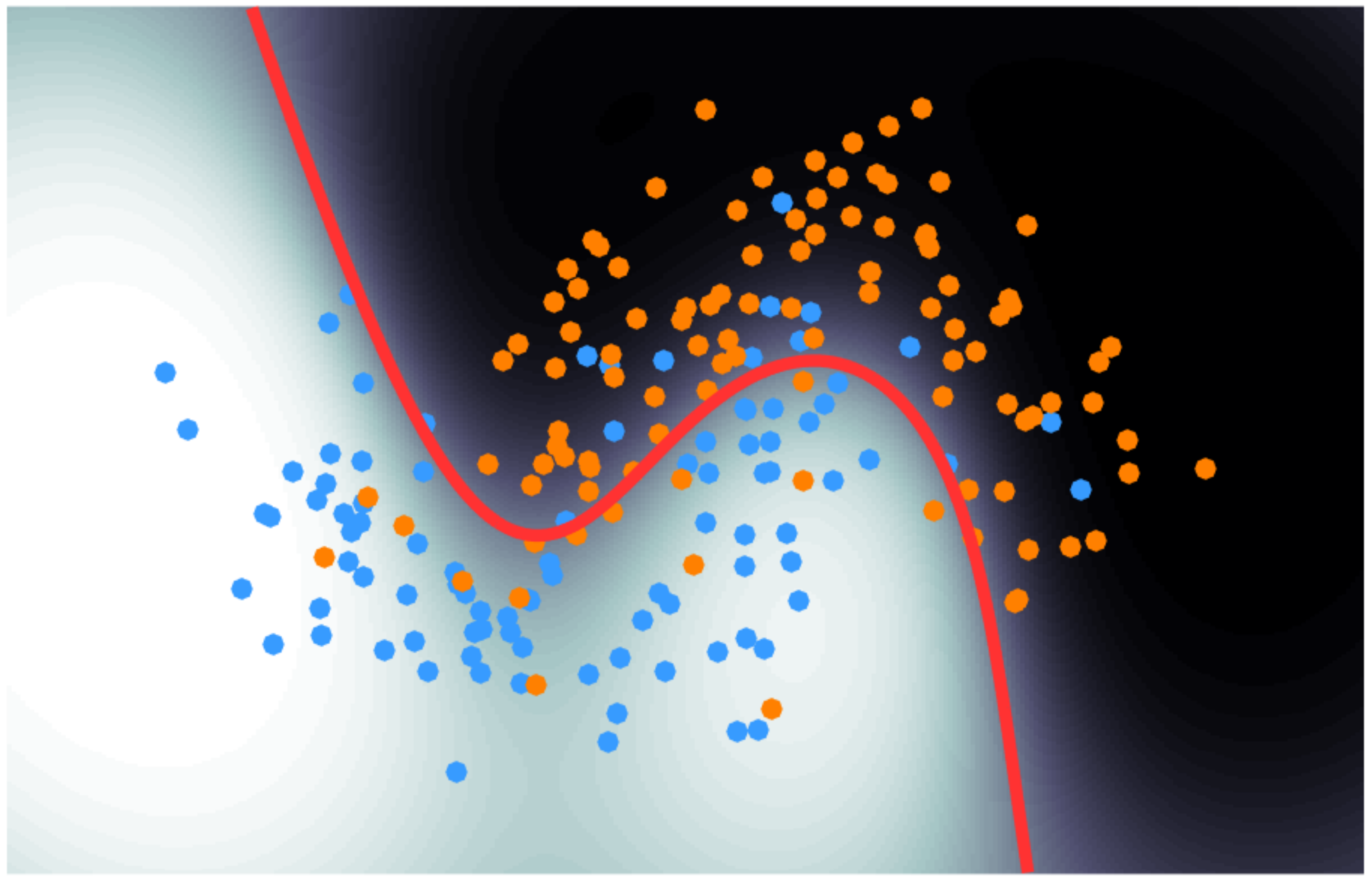} &
  \includegraphics[height=.18\textwidth,angle=90]{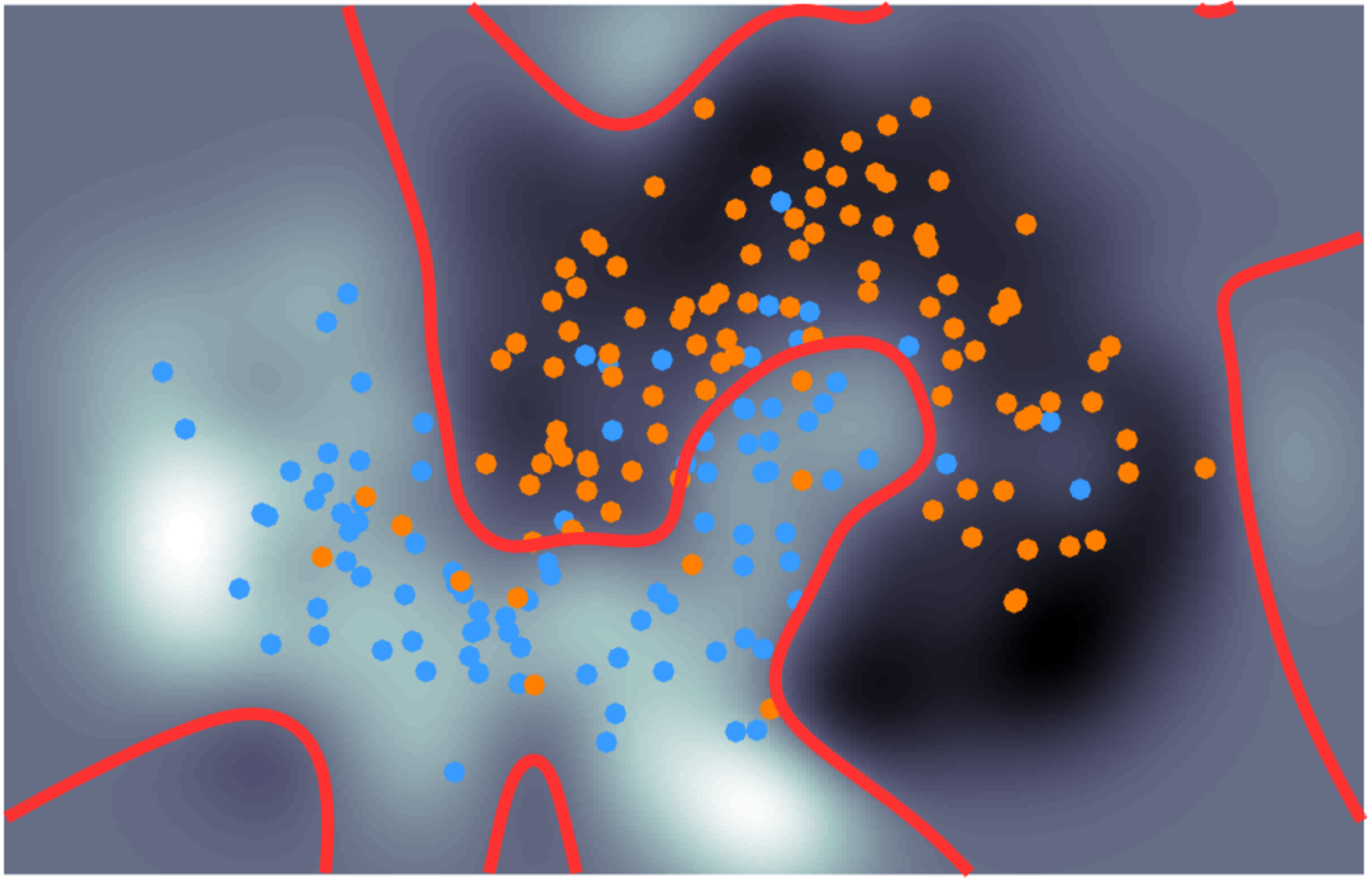}\\
  (a) & (b) & (c) & (d)
  \end{tabular}
\vspace{-.1in}
  \caption{Comparison of classifiers with different regularizers. For ease of exposition, we only draw training data (blue and orange markers) and the classification boundary (red). (a): our method achieves structural simplicity without over-smoothing the classifier boundary. A standard classifier (\eg, kernel method using the same $\sigma$) could (b) overfit, or (c) overly smooth the classification boundary and reduce overall accuracy. 
  (d): The output of the STOA method based on geometrical simplicity \citep{bai2016differential} also smooths the classifier globally.}
\label{fig:teaser}
\vspace{-.2in}
\end{figure*}

Regularization plays a crucial role in supervised learning. 
A successfully regularized model strikes a balance between a perfect description of the training data and the ability to generalize to unseen data. 
A common intuition for the design of regularzers is the Occam's razor principle, where a regularizer enforces certain simplicity of the model in order to avoid overfitting.
Classic regularization techniques include functional norms such as $L_1$ \citep{krishnapuram2005learning}, $L_2$ (Tikhonov) \citep{ng2004feature} and RKHS norms \citep{scholkopf2002learning}. Such norms produce a model with relatively less flexibility and thus is less likely to overfit. 

A particularly interesting category of methods is inspired by the geometry.
These methods design new penalty terms to enforce a geometric simplicity of the classifier. Some methods stipulate that similar data should have similar score according to the classifier, and enforce the smoothness of the classifier function 
\citep{Belkin2006,Zhou2005,bai2016differential}. Others directly pursue a simple geometry of the classifier boundary, \ie, the submanifold separating different classes \citep{CaiSowmya2007,VarshneyWillsky2010,Lin2012,Lin2015}. 
These geometry-based regularizers are intuitive and have been shown to be useful in many supervised and semi-supervised learning settings. 
However, regularizing total smoothness of the classifier (or that of the classification boundary) is not always flexible enough to balance the tug of war between overfitting and overall accuracy. 
The key issue is that these measurement are usually structure agnostic.
For example, in Figure \ref{fig:teaser}, a classifier may either overfit (as in (b)), or becomes too smooth and lose overall accuracy (as in (c)). 

\begin{figure*}[hbtp]
\vspace{-.1in}
  \centering
  \begin{tabular}{ccc}
      \includegraphics[width=.3\textwidth]{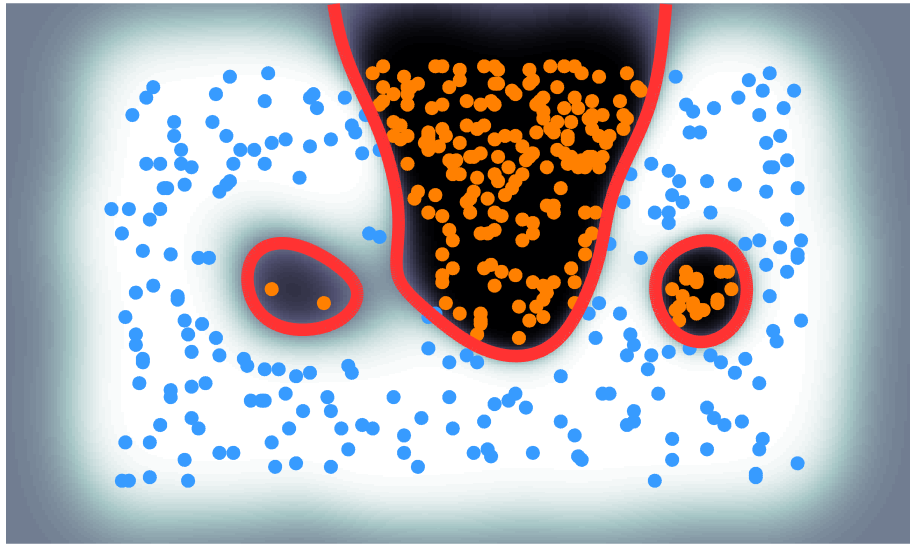} &
      \hspace{-.1in}
      \includegraphics[width=.3\textwidth]{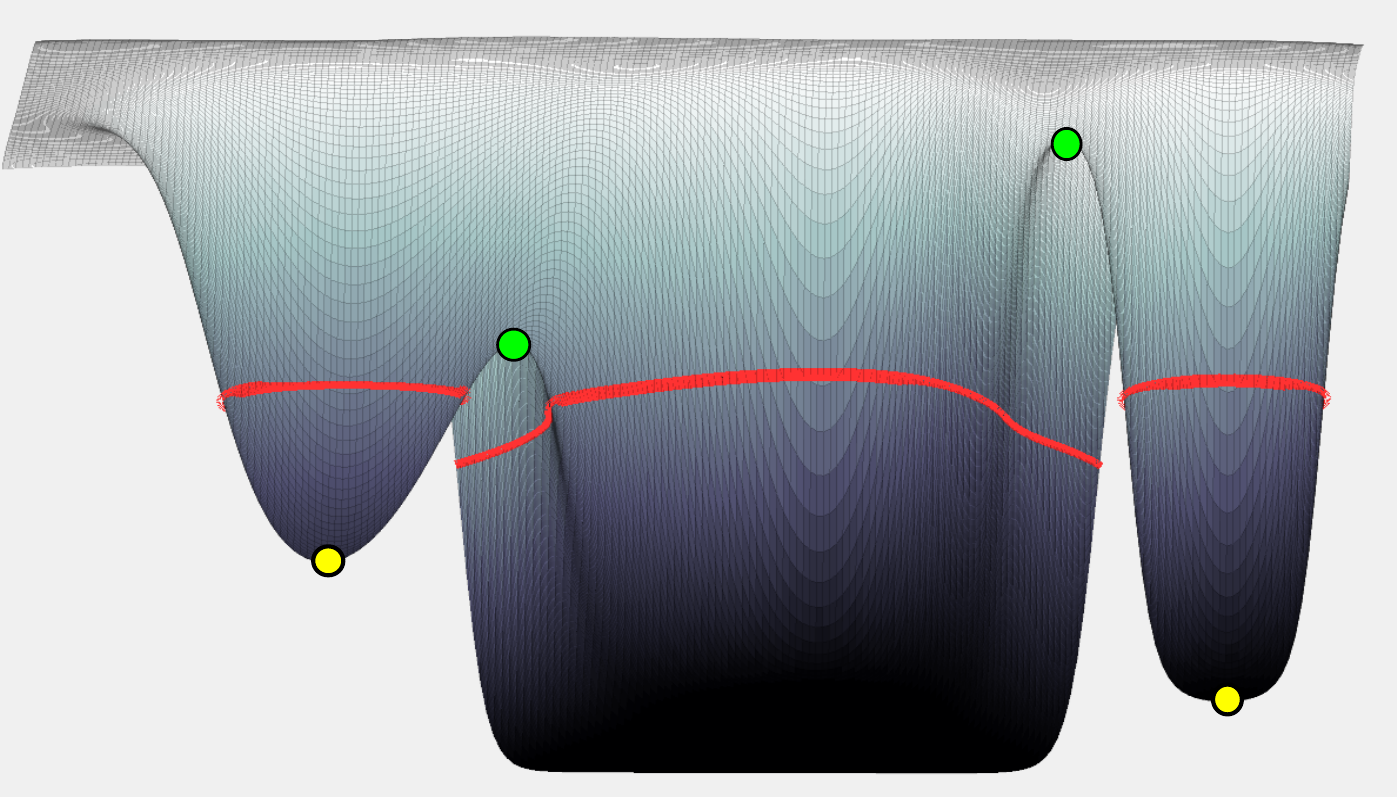} &
      \hspace{-.1in}
  \includegraphics[width=.28\textwidth]{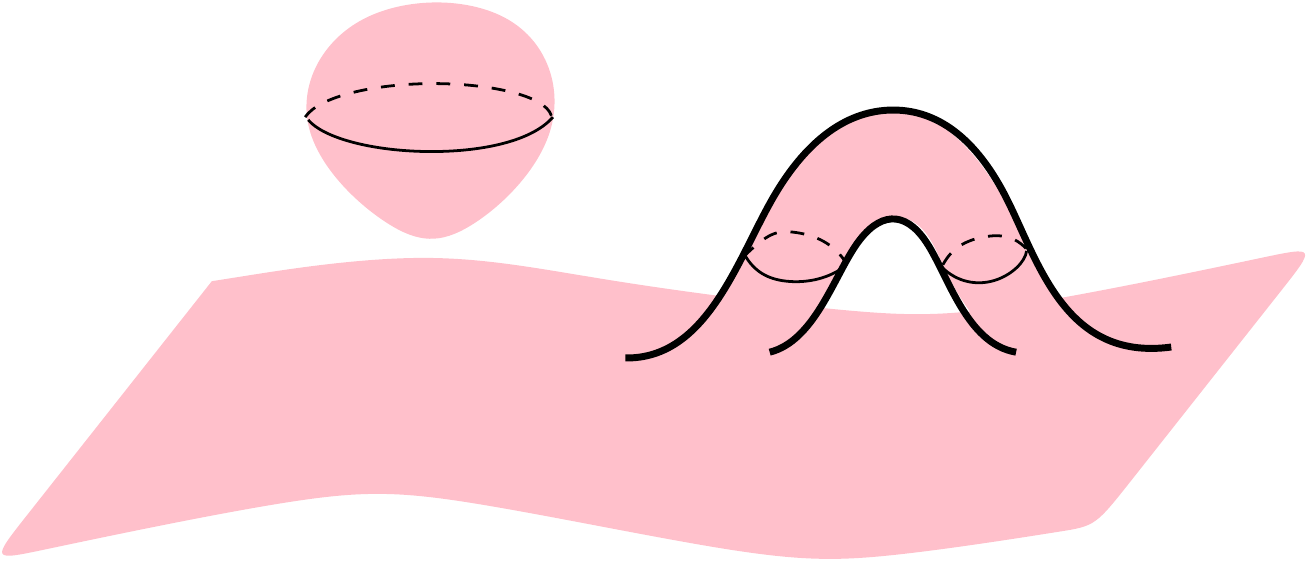}\\
  (a) & (b) & (c)
  \end{tabular}
\vspace{-.1in}
  \caption{(a): The classifier boundary (red curves) has two additional connected components with similar geometry. But the left one is in fact less important \wrt the classifier as shown in (b), where the graph of the classifier function is shown (i.e, using $f$ as the elevation function). The left valley is easier to be removed from the 0-value level set by perturbation. (c): The classifier boundary can have different types (and dimensional) topological structures, \eg, connected components, handles, voids, \etc. 
  }
\label{fig:topology}
\vspace{-.1in}
\end{figure*}

In this paper, we propose a new direction to regularize the ``simplicity'' of a classifier -- Instead of using geometry such as total curvature / smoothness, we directly enforce the ``simplicity'' of the classification boundary, by regularizing over its \emph{topological complexity}. (Here, we take a similar functional view as Bai \etal \citeyearpar{bai2016differential} and consider the classifier boundary as the $0$-valued level set of the classifier function $f(x)$; see Figure \ref{fig:topology} for an example.) 
Furthermore, our measurement of topological complexity incorporates the \emph{importance} of topological structures, \eg, connected components, handles, in a meaningful manner, and provides a direct control over spurious topological structures. 
This new structural simplicity can be combined with other regularizing terms (say geometry-based ones or functional norms) to train a better classifier. See Figure \ref{fig:teaser} (a) for an example, where the classifier computed with topological regularization 
achieves a better balance between overfitting and classification accuracy. 

To design a good topological regularizer, there are two key challenges.
First, we want to measure and incorporate the significance of different topological structures. For example, in Figure \ref{fig:topology} (a), we observe three connected components in the classification boundary (red). The ``importance'' of the two smaller components (loops) are different despite their similar geometry. The component on the left exists only due to a few training data and thus are much less robust to noise than the one on the right.
%
%
Leveraging several recent developments in the field of computational topology \citep{edelsbrunner2000topological,bendich2010computing,bendich2013homology}, we quantify such ``robustness'' $\rho(c)$ of each topological structure $c$ and define our topological penalty as the sum of the squared robustness $\LT(f) = \sum \rho(c)^2$ over all topological structures from the classification boundary. 

A bigger challenge is to compute the gradient of the proposed topological penalty function.
In particular, the penalty function crucially depends on locations and values of \emph{critical points} (\eg, extrema and saddles) of the classifier function. But there are no closed form solutions for these critical points.
To address this issue, we propose to discretize the domain and use a piecewise linear approximation of the classifier function as a surrogate function.
We prove in Section \ref{sec:method} that by restricting to such a surrogate function, the topological penalty is differentiable almost everywhere.
We propose an efficient algorithm to compute the gradient and optimize the topological penalty.
We apply the new regularizer to a kernel logistic regression model and show in Section \ref{sec:exp} how it outperforms others on various synthetic and real-world datasets.

In summary, our contributions are as follows: 
\begin{itemize}[topsep=1pt, partopsep=1pt,itemsep=0pt]
\item We propose the novel view of regularizing the topological complexity of a classifier, and present the first work on developing such a topological penalty function; 
\item We propose a method to compute the gradient of our topological penalty. 
By restricting to a surrogate piecewise linear approximation of the classifier function, we prove the gradient exists almost everywhere and is tractable; 
\item We instantiate our topological regularizer on a kernel classifier.  
We provide experimental evidence of the effectiveness of the proposed method on several synthetic and real-world datasets.
\end{itemize}

Our novel topological penalty function and the novel gradient computation method open the door for a seamless coupling of topological information and learning methods.
For computational efficiency,
in this paper, we focus on the simplest type of topological structures, \ie, connected components.
The framework can be extended to more sophisticated topological structures, \eg, handles, voids, \etc.

%

\myparagraph{Related work.}
The topological summary called persistence diagram/barcode (will be defined in the supplemental material) can capture the global structural information of the data \emph{in a multiscale manner}. It has been used in unsupervised learning, \eg, clustering \citep{chazal2013persistence,ni2017composing}. In supervised setting, topological information has been used as powerful features. The major challenge is the metric between such topological summaries of different data is not standard Euclidean. Adams \etal \citeyearpar{adams2017persistence} proposed to directly vectorize such information. Bubenik \citeyearpar{bubenik2015statistical} proposed to map the topological summary into a Banach space so that statistical reasoning can be carried out \citep{chazal2014convergence}. To fully leverage the topological information, various kernels \citep{reininghaus2015stable,kwitt2015statistical,kusano2016persistence,carriere2017sliced,zhu2016stochastic} have been proposed to approximate their distance.
Hofer \etal \citeyearpar{hofer2017deep} proposed to use the topological information as input for deep convolutional neural network.
Perhaps the closest to us are \citep{varshney2015persistent,RVM18}, which compute the topological information of the classification boundary. 
All these methods use topological information as an observation/feature of the data. \superemph{To the best of our knowledge, our method is the first to leverage the topological information as a prior for training the classifier.}

In the context of computer vision, topological information has been incorporated as constraints in discrete optimization. Connectivity constraints can be used to improve the image segmentation quality, especially when the objects of interest are in elongated shapes. However in general, topological constraints, although intuitive, are highly complex and too expensive to be fully enforced in the optimization procedure \citep{vicente2008graph,nowozin2009global}. One has to resort to various approximation schemes \citep{zeng2008topology,chen2011enforcing,stuhmer2013tree,oswald2014generalized}.

\section{Level Set, Topology, and Robustness}
\label{sec:background}

To illustrate the main ideas and concepts, we first focus on the binary classification problem\footnote{For the multilabel classification, we will use multiple one-vs-all binary classifiers (see Section \ref{sec:method}).} with a $D$-dimensional feature space, $\calX \subset \R^D$. 
\Wlg (without loss of generality), we assume $\calX$ is a $D$-dimensional hypercube, and thus is compact and simply connected.
A classifier function is a smooth scalar function, \mbox{$f:\calX\rightarrow \R$}, and the prediction for any training/testing data $x\in \calX$ is $\sgn(f(x))$. 
We are interested in describing the topology and geometry of the \emph{classification boundary} of $f$, \ie, the boundary between the positive and negative classification regions.
Formally, the boundary is the \emph{level set} of $f$ at value zero, \ie, the set of all points with function value zero $$S_f = f^{-1}(0) = \{ x\in \calX | f(x) = 0\}.$$
\Wlg, we assume $S_f$ is a $(D-1)$-dimensional manifold, possibly with multiple connected components\footnote{The degenerate case happens if $S_f$ passes through critical points, \eg, saddles, minima, or maxima.}. 
In Figure \ref{fig:topology}(a), the red curves represent the boundary $S_f$, which is a one-dimensional manifold consisting of three connected components (one U-shaped open curve and two closed loops).
Please note that a level set has been used extensively in the image segmentation tasks \citep{osher2006level,szeliski2010computer}. 

\begin{figure}[btp!]
  \vspace{-.1in}
  \centering
  \begin{tabular}{cccc}
      \hspace{-.15in}
      \includegraphics[height=.12\textwidth]{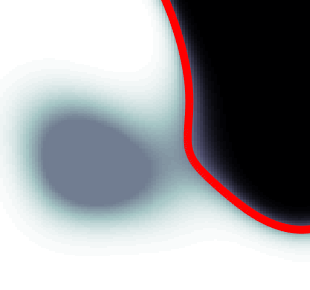} &
      \hspace{-.15in}
      \includegraphics[height=.12\textwidth]{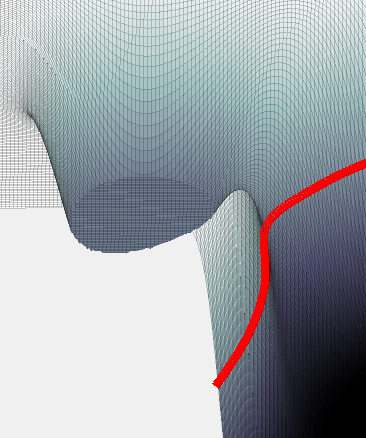} &
      \hspace{-.15in}
      \includegraphics[height=.12\textwidth]{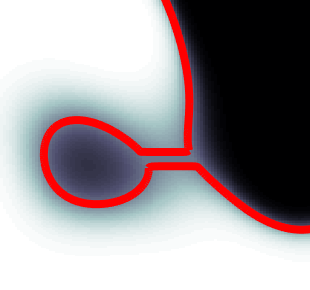} &
      \hspace{-.15in}
      \includegraphics[height=.12\textwidth]{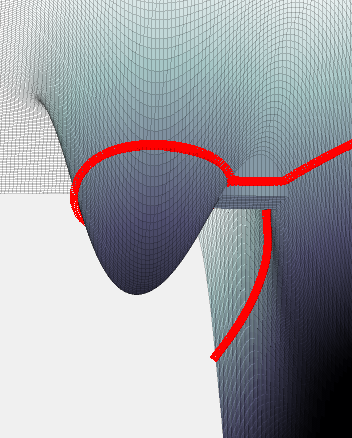} \\

 (a) & (b) & (c) & (d) 
  \end{tabular}
  \vspace{-.1in}
  \caption{Two options to eliminate the left loop in Figure \ref{fig:topology}(a).
  Option 1: increase values of all points inside the loop so the loop disappears completely. (a): zoom-in of the new function. (b): the graph of the new function. 
Option 2: decrease values along a path through the saddle so the loop is merged with the U-shaped curve. (c): zoom-in of the new function, (d): the graph of the new function. 
}
\label{fig:fixing}
  \vspace{-.2in}
\end{figure}

For ease of exposition, we only focus on the simplest type of topological structures, \ie, connected components.
For the rest of the paper, unless specifically noted, we will use ``connected components'' and ``topological structures'' interchangeably.
Classification boundaries of higher dimension may have other types of topological structures, \eg, handles, voids, \etc. See Figure \ref{fig:topology}(c) for the boundary of a 3D classifier.
Our method can be extended to these structures, as the mathematical underpinning is well understood \citep{edelsbrunner2010computational}.



\myparagraph{Robustness.}
Our goal is to use the topological regularizer to simplify the topology of a classifier boundary. To achieve so, we need a way to rank the significance of different topological structures. The measure should be based on the underlying classifier function. To illustrate the intuition, recall the example in Figure \ref{fig:topology}(a). To rank the three connected components of the classifier boundary $S_f$, simply inspecting the geometry is insufficient. The two loops have similar size. However, the left loop is less stable as it is caused by only a few training samples (two positive samples inside the loop and two negative samples between the loop and the U-shaped curve). Instead, the difference between the two loops can be observed by studying the graph of the function (Figure \ref{fig:topology}(b)). 
Compared to the right loop, the basin inside the left loop is shallower and the valley nearby is closer to the sea level (zero). 

We propose to measure the significance of a component of interest, $c$, as the minimal amount of necessary perturbation the underlying classifier $f$ needs in order to ``shrug off'' $c$ from the zero-valued level set. Formally, we define the \emph{robustness} for each connected component of the boundary $S_f$ as follows.
\begin{definition}[Robustness]
    The robustness of $c$ is $\rho(c) = \min_{\hat{f}} \dist(f,\hat{f})$, so that $c$ is not a connected component of the boundary of the perturbed function $\hat{f}$. 
The distance between $f$ and its perturbed version $\hat{f}$ is via the $L_\infty$ norm, \ie, $\dist(f,\hat{f}) = \max_{x\in \calX} |f(x)-\hat{f}(x)|$. 
\end{definition}

In the example of Figure \ref{fig:topology}, there are two options to perturb $f$ and so the left loop can be eliminated:

\superemph{Option 1.} Remove the left loop completely by increasing the function value of all points within it to $+\epsilon$, where $\epsilon$ is an infinitesimally small positive value. For the new function $g$, the zero-valued level set only consists of the U-shaped curve and the right loop. See Figure \ref{fig:fixing}(a) and (b) for a zoomed-in view of $g$ and its graph. In such case, the expense $\dist(f,g)$ is simply $\epsilon$ plus the depth of the basin, \ie, the absolute function value of the local minimum inside the left loop (left yellow marker in Figure \ref{fig:topology}(b)). The actual expense is $\dist(f,g) = |-0.55|+\epsilon = 0.55+\epsilon$.

\superemph{Option 2.} Merge the left loop with the U-shaped curve by 
finding a path connecting them and lowering the function values along the path to $-\epsilon$. There are many paths achieving the goal. But note that all paths connecting them have to go at least as high as the nearby saddle point (left green marker in Figure \ref{fig:topology}(b)). Therefore, we choose a path passing the saddle point and has the saddle as the highest point. By changing function values along the path to $-\epsilon$, we get the new function $h$.
    In the zero-valued level set of $h$, $S_h$, the left loop is merged with the U-shaped curve via a ``pipe''. See Figure \ref{fig:fixing}(c) and (d) for a zoomed-in view of $h$ and its graph.
    In this case, the expense $\dist(f,h)$ is $\epsilon$ plus the highest height of the path, namely, the function value of the saddle point. $\dist(f,h) = 0.21+\epsilon$. 

    To optimize the cost to remove the left loop, we choose the second option. The corresponding expense gives us the robustness of this left component $c$, $\rho(c)=0.21+\epsilon$. In practice and for the rest of the paper, we drop $\epsilon$ for convenience.
Note for the right loop, its robustness is much higher as values of the associated critical points are further away from the value zero. In fact, its robustness is $0.83$.

%
\begin{figure*}[hbtp]
\vspace{-.1in}
  \centering
  \begin{tabular}{cc}
  \includegraphics[height=.25\textwidth]{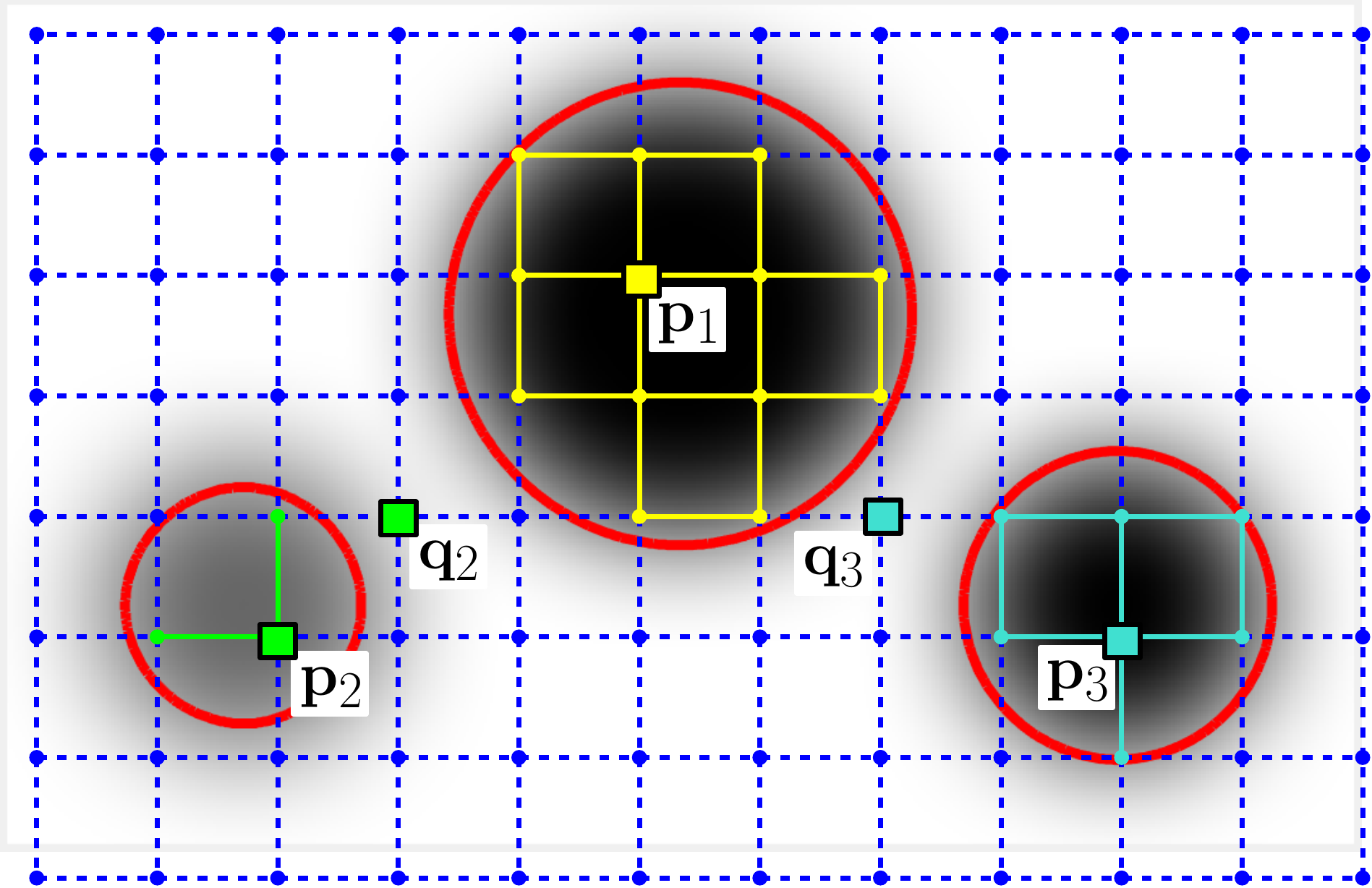} &
  \includegraphics[height=.25\textwidth]{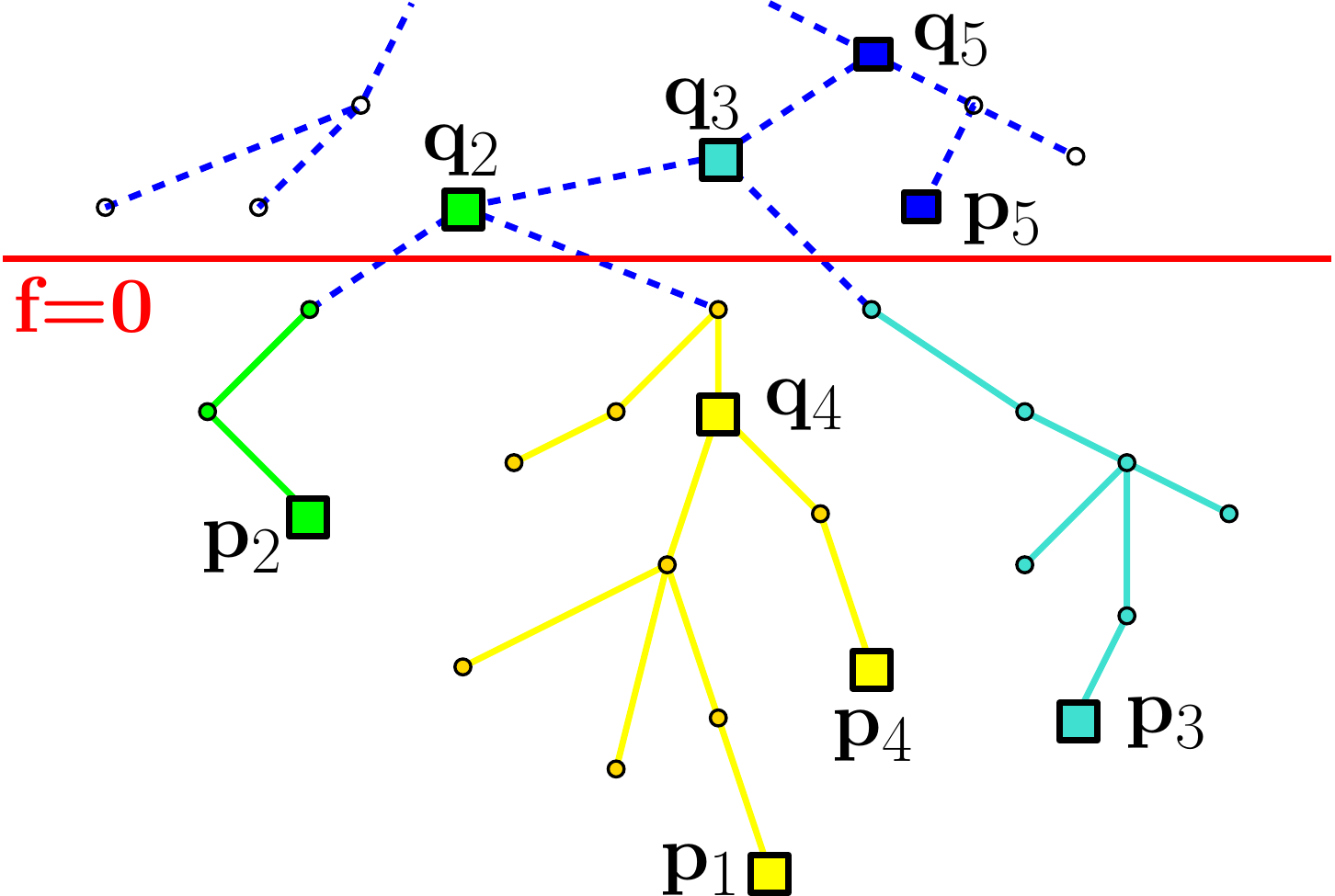} \\
  (a) & (b)
  \end{tabular}
\vspace{-.1in}
\caption{Illustration of the algorithm. (a) grid graph on a given function $f$. Red curves are the classification boundary (zero-valued level set). (b) the corresponding merging tree. the green tree is created at its minimum $p_2$ and is merged to the yellow tree at the saddle $q_2$. The turquoise subtree is created at its minimum $p_3$ and is merged to the yellow tree at saddle $q_3$. Their corresponding pairings are $(p_2,q_2)$ and $(p_3,q_3)$, respectively. Some other subtrees have their pairings in $\widehat{\Pi}$ but not in $\Pi_f$. They either are created after zero (\eg, $(p_5,q_5)$) or are killed before zero (\eg, $(p_4,q_4)$).}
\label{fig:alg}
\vspace{-.2in}
\end{figure*}

In this example, we observe that the robustness of a component crucially depends on the function values of two critical points, a minimum $p$ and a saddle point $q$. 
This is not a coincidence: this pairing ($p, q$) is in fact a so-called \emph{persistence pairing} computed based on the theory of persistent homology \citep{ELZ02,CSM09,bendich2013homology}. We skip the description here, and refer the readers to the supplemental material for some details. 
We provide the following theorem, and the proof can be found in the supplemental material. 
\begin{theorem}\label{thm:perpairs}
Let $f: \calX \to \R$ be a Morse function defined on a $D$-dimensional hypercube $\X \subset \R^D$. 
Then, there is a set of pairings of critical points of $f$, $\Pi = \{(p, q) \}$, such that there is a one-to-one correspondence between the set of connected components of the boundary $S_f$ and pairings in $\Pi$. In particular, under this bijection, suppose a component $c$ corresponds to a pair of critical points $(p_c, q_c)$. Then 
its robustness $\rho(c) = \min \{|f(p_c)|, |f(q_c)|\}$. 

Furthermore, $\Pi$ can be computed by computing the so-called 0-th dimensional persistent homology induced by sublevel set filtration \wrt function $f$ and \wrt function $-f$, respectively. 
\end{theorem}

The second part of the theorem outlines an efficient algorithm for computing robustness. Below we present the details of the algorithm, and discuss its complexity.

\myparagraph{Algorithm.} 
According to the theorem, we need to compute 0-dimensional persistent homology for both $f$ and $-f$ in order to collect all necessary pairings of critical points.
We first describe now how to compute the portion of pairings in $\Pi$ coming from the 0-th persistent homology \wrt $f$, \ie, $\Pi_f\subseteq \Pi$. A symmetric procedure will then be performed for $-f$ to compute $\Pi_{-f}\subseteq \Pi$. 
For computation purpose, we need a discretization of the domain $\calX$ and the classifier function $f$ are evaluated at vertices of this discretization. 
For low-dimensional feature space, \eg, 2D, we take a uniform sampling of $\calX$ and evaluate the function at all samples. A grid graph $G = (V, E)$ is built using these samples as vertices, and we can evaluate the function value of $f$ at all vertices $V$. See Figure \ref{fig:alg}(a). 

Next, we build a merging tree as well as a collection of pairings $\widehat{\Pi}$ as follows. We sort all vertices according to their function values $V = \{v_1, \ldots, v_n\}$. Add these vertices one-by-one according to the order (from small to large). At any moment $i$, we maintain the spanning forest for all the vertices $V_i = \{v_1, \ldots, v_i \}$ that we already swept. 
Furthermore, each tree in the spanning forest is represented by the global minimum $p_m$ in this tree. When two trees  $T_1$ and $T_2$ (associated with global minima $p_1$ and $p_2$, respectively) merge upon the processing of node $v_s$, then the resulting tree is represented by the lower of the two minima $p_1$ and $p_2$, say $p_1$, and we add the pairing $(p_2, v_2)$ to $\widehat{\Pi}$. Intuitively, the tree $T_2$ is created when we sweep past $p_2$, and is ``killed'' (merged to an ``older'' tree $T_1$ created at $p_1$ with $f(p_1) \le f(p_2)$). 

After all vertices are added, the merging tree $T_f$ is constructed. See Figure \ref{fig:alg}(b), where the height of tree nodes corresponds to their function values. 
The entire process can be done by a standard union-find data structure with a slight modification to maintain the minimum of each set (tree) in $O(m\alpha(n))$ time once the vertices are sorted \citep{edelsbrunner2010computational}, where $m$ is the number of edges in grid graph $G$, $\alpha(n)$ is the inverse Ackermann function.

This merging tree and the set of pairings $\widehat{\Pi}$ encode all the information we need. If we cut the tree at level zero, the remaining subtrees one-to-one correspond to different connected components of the boundary. 
Indeed, now set $\Pi_f := \{ (p, q) \in \widehat{\Pi} \mid f(p) < 0 < f(q) \}$: each of the pair $(p, q)$ in $\Pi_f$ corresponds to one component in the separation boundary $S_f$; the tree containing $p$ it was created at $f(p)$ before $0$ but merged at $f(q)$ after $0$. In Figure \ref{fig:alg}(b), the green, yellow, and turquoise subtrees correspond to the three connected components of the boundary of the function in Figure \ref{fig:alg}(a). Notice that both the green and the turquoise trees/components have their corresponding pairings in $\Pi_f$, $(p_2,q_2)$ and $(p_3,q_3)$, respectively. However, the yellow subtree/component does not have a pairing in $\Pi_f$ as it is not merged into anyone. The pairing for the yellow component will be $(v_1,v_n)$, in which $v_1=p_1$ is the global minimum and $v_n$ is the global maximum. 

We perform the same procedure but for function $-f$, and collect $\Pi_{-f}$. Finally, (via the proof of Theorem \ref{thm:perpairs} in the supplemental material), the set of critical pairs $\Pi$ corresponding to the set of components in $S_f$ is $\Pi = \Pi_f \cup \Pi_{-f} \cup \{ (v_1, v_n) \}$. 
The complexity of the algorithm is $O(n \log n + m \alpha(n))$. The first term comes from the sorting of all vertices in $V$. The second term is due to the merge-tree building using union-find. 

The grid-graph discretization is only feasible for low-dimensional feature space. For high dimension, we compute a discretization of the feature space $\calX$ using a k-nearest-neighbor graph (KNN) $G' = (V', E)$: Nodes of this graph are all training data points. Thus the extracted critical points are only training data points.
We then perform the same procedure to compute $\Pi$ as described above using this $G'$. 
Our choice is justified by experimental evidences that the KNN graph provides sufficient approximation for the topological computation in the context of clustering \citep{chazal2013persistence,ni2017composing}.

\section{Topological Penalty and Gradient}
\label{sec:method}
Based on the robustness measure, we will introduce our topological penalty below. 
To use it in the learning context, a crucial step is to derive the gradient.
However, the mapping from input data to persistence pairings ($\Pi$ in Theorem \ref{thm:perpairs}) is highly non-linear without an explicit analytical representation. Hence it is not clear how to compute the gradient of a topological penalty function in its original format. 
Our key insight is that, if we approximate the classifier function by a piecewise-linear function, then we can derive gradients for the penalty function, and perform gradient-descent optimization.
Our topological penalty is implemented on a kernel logistic regression classifier, and we also show how to extend it to multilabel settings. 

Given a data set $\calD=\{(x_n,t_n)\mid n = 1,\ldots,N\}$ and a classifier $f(x,w)$ parameterized by $w$, we define the objective function to optimize as the weighted sum of the per-data loss and our topological penalty. 
\begin{align} 
    L(f,\calD) &= \sum_{(x,t)\in \calD} \ell(f(x,w),t) + \lambda \LT(f(\cdot,w)), 
\label{eq:loss}
\end{align}
in which $\lambda$ is the weight of the topological penalty, $\LT$.
And $\ell(f(x,w),t)$ is the standard per-data loss, \eg, cross-entropy loss, quadratic loss or hinge loss.

Our topological penalty, $\LT$, aims to eliminating the connected components of the classifier boundary. 
In the example of Figure \ref{fig:topology}(a), it may help eliminating both the left and the right loops, but leaving the U-shaped curve alone as it is the most robust one.
Recall each topological structure of the classification boundary, $c$, is associated with two critical points $p_c$ and $q_c$, and its robustness $\rho(c) = \min\{|f(p_c,w)|, |f(q_c,w)|\}$.
We define the topological penalty in Equation \eqref{eq:loss} as the sum of squared robustness, formally,
%
\vspace{-.05in}
$$ 
\LT(f) = \sum_{c\in \calC(S_f)} \rho(c)^2. 
\vspace{-.05in}
$$
Here $\calC(S_f)$ is the set of all connected components of $S_f$ except for the most robust one. In Figure \ref{fig:topology}(a), $\calC(S_f)$ only consists of the left and the right loops. We do not include the most robust component, as there should be at least one component left in the classifier boundary. 

\myparagraph{Gradient.}
A crucial yet challenging task is to compute the gradient of such topological penalty.
In fact, there has not been any gradient computation for topology-inspired measurement. A major challenge is the lack of a closed form solution for the critical points of any non-trivial function. Previous results show that even a simple mixture of isotropic Gaussians can have exponentially many critical points \citep{edelsbrunner2013add,carreira2003number}.

In this paper, we propose a solution that circumvents the direct computation of critical points in the continuous domain. The key idea is to use a piecewise linear approximation of the classifier function. Recall we discretize the feature space into a grid graph, $G=(V,E)$, and only evaluate classifier function values at a finite set of locations/points. Now consider the piecewise linear function $\hat{f}$ which agrees with $f$ at all sample points in $V$, but linearly interpolates along edges. We show that restricting to such piecewise linear functions, the gradient of $\LT$ is indeed computable.
\begin{theorem}
    Using the piecewise linear approximation $\hat{f}$, the topological penalty $\LT(\hat{f}(\cdot,w))$ is differentiable almost everywhere. 
\end{theorem}
\proof
For the piecewise linear approximate $\hat{f}$, all critical points have to come from the set of vertices $V$ of the discretization. Their pairing and correspondence to the connected components can be directly computed using the algorithm in Section \ref{sec:background}.

We first assume $\hat{f}$ has unique function values at all points in $V$, i.e, $\hat{f}(u)\neq \hat{f}(v)$, $\forall u,v\in V$. 
Let $\Delta$ be the lowerbound of the difference between the absolute function values of elements in $V$, as well as the absolute function values of all vertices:
\mbox{$\Delta = \min\{\min_{u,v\in V, u\neq v} \vert\vert\hat{f}(u)\vert - \vert\hat{f}(v)\vert \vert, \min_{u\in V} \vert\hat{f}(u)\vert\}$}
To prove our theorem, we show that there exists a small neighborhood of the function $\hat{f}$, so that for any function in this neighborhood, the critical points and their pairings remain unchanged. 
To see this, we note that the any function in such neighborhood of $\hat{f}$ is also piecewise linear functions realized on the same graph $G$. We define the neighborhood to be a radius $\Delta/2$ open ball in terms of $L_\infty$ norm, formally, 
$\calF = \{g \mid \vert\hat{f}(v) - g(v)\vert < \Delta/2, \forall v\in V\}.$ 
For any $g \in \calF$, $\forall u,v\in V$, we have the following three facts:
\begin{description}[topsep=0pt, itemsep = 0pt]
    \item[Fact 1.] $\hat{f}(u) < \hat{f}(v)$ if and only if $g(u) < g(v)$.
    \item[Fact 2.] $\hat{f}(u) < 0$ if and only if $g(u) < 0$, and $\hat{f}(u) > 0$ if and only if $g(u) > 0$ 
    \item[Fact 3.] $\vert\hat{f}(u) \vert< \vert\hat{f}(v)\vert$ if and only if $\vert g(u) \vert< \vert g(v)\vert$.
\end{description}
The first two facts guarantee that the ordering of elements in $V$ induced by their function values are the same for $g$ and $\hat{f}$. 
Consequently, the filtration of all elements of $G$ induced by $g$ and $\hat{f}$ are the same. By definition of persistent homology, the persistence pairs (of critical points) are identical for $g$ and $\hat{f}$. In other words, the pair associated with each connected component $c$, $(p_c,q_c)$ are the same for both $g$ and $\hat{f}$.

Furthermore, the third condition guarantees that for each $c$, $g(p_c) < g(q_c)$ if and only if $\hat{f}(p_c) < \hat{f}(q_c)$. If for $\hat{f}$, $p_c$ is the critical point that accounts for the robustness, \ie, $\rho(c) = \hat{f}(p_c)$, then $p_c$ also accounts for $\rho(c)$ for function $g$. Denote by $p_c^\ast$ as such critical point. 
$
p_c^\ast = \argmin_{p\in\{p_c,q_c\}} \{\vert\hat{f}(p_c)\vert,\vert \hat{f}(q_c) \vert\}
=\argmin_{p\in\{p_c,q_c\}} \{\vert g(p_c)\vert,\vert g(q_c) \vert\}.
$
Thus \mbox{$\rho(c) =\vert \hat{f}(p_c^\ast) \vert$}, in which the critical point $p_c^\ast$ remains a constant for any $g$ within the small neighborhood of $\hat{f}$. 

With constant $p_c^\ast$'s, 
and knowing that $\hat{f}$ and $f$ agree at all elements of $V$, the gradient is straightforward
\begin{multline*}
    \nabla_w \LT = \sum_{c\in \calC(S_{\hat{f}})} \nabla_w (\rho(c)^2) 
                 =  \sum_{c\in \calC(S_{\hat{f}})} 2 f(p_c^\ast,w) \frac{\partial f(p_c^\ast)}{\partial w}.
\end{multline*}
%
Note that this gradient is intractable without the surrogate piecewise linear function $\hat{f}$; for the original classifier $f$, $p_c^\ast$ changes according to $w$ in a complex manner.

Finally, we note that it is possible that elements in $V$ may have the save function values or the same absolute function values. In such cases, $p_c^\ast$ may not be uniquely defined and the gradient does not exist. However, these events constitute a measure zero subspace of functions and do not happen generically. In other words, $\LT(\hat{f})$ is a piecewise smooth loss function over the space of all piecewise linear functions. It is differentiable \emph{almost everywhere}. 
\qed

%
%
%
\myparagraph{Intuition of the gradient.}
During the optimization process, we take the opposite direction of the gradient, \ie, $-\nabla_w\LT$. For each component $c\in \calC(S_f)$, taking the direction $-\nabla_w(\rho(c)^2)$ is essentially pushing the function value of the critical point $p_c^\ast$ closer to zero. In the example of Figure \ref{fig:topology}, for the left loop, the gradient decent will push the function value of the saddle point (left green marker) closer to zero, effectively dragging the path down as in Figure \ref{fig:fixing}(c) and (d). If it is the case when $p_c^\ast$ is the minimum, the gradient descent will increase the function value of the minimum, effectively filling the basin as in Figure \ref{fig:fixing}(a) and (b).
\myparagraph{Instantiating on kernel machines.}
In principle, our topological penalty can be incorporated with any classifier. Here, we combine it with a kernel logistic regression classifier to demonstrate its advantage. We first present details for a binary classifier. We will extend it to multilabel settings. 
For convenience, we abuse the notation, drop $\hat{f}$ and only use $f$.

In a kernel logistic regression, the prediction function is 
$f(x,w) = g\left(\phi(x)^Tw\right) = 1/\left(1+\exp\left(-\phi(x)^Tw\right)\right)$.
The $N$-dim feature $\phi(x) = (k(x,x_1),\cdots,k(x,x_N))^T$ consists of the Gaussian kernel distance between $x$ and the $N$ training data.
The per-data loss $\ell(f(x,w),t)$ is the standard cross-entropy loss and its gradient can be found in a standard textbook \citep{bishop2006pattern}. 

Next we derive the gradient for the topological penalty. First we need to modify the classifier slightly. Notice the range of $f$ is between zero and one, and the prediction is $\sgn(f-0.5)$. To fit our setting in which the zero-valued level set is the classification boundary, we use a new function $\tilde{f} = f-0.5$ as the input for the topological penalty.
The gradient is
\begin{align*}
    & \nabla_w \LT   = \sum_{c\in \calC(S_f)} 2\tilde{f}(p_c^\ast,w)\frac{\partial \tilde{f}(p_c^\ast)}{\partial w} \nonumber \\
    & = \sum_{c\in \calC(S_f)} \left(-2f(p_c^\ast,w)^3 + 3f(p_c^\ast,w)^2-f(p_c^\ast,w)\right) \phi(p_c^\ast)
\end{align*}
Our overall algorithm repeatedly computes the gradient of the objective function (gradient of cross-entropy loss and gradient of topological penalty), and update the parameters $w$ accordingly, until it converges.
At each iteration, to compute the gradient of the topological penalty, we compute the 
critical point $p_c^\ast$'s for all connected components, using the algorithm introduced in Section \ref{sec:background}.


\myparagraph{Multilabel settings.}
For multilabel classification with $K$ classes, we use the multinomial logistic regression classifier $f^k(x,W), k = 1..K$ with parameters $W = (w^1,\cdot,w^K)$.
The per-data loss is again the standard cross-entropy loss. For the topological penalty, we create $K$ different scalar functions $\psi^k(x,W) = \max_{t\neq k} f^t(x,W) - f^k(x,W)$. If $\psi^k(x,W) < 0$, we classify $x$ as label $k$. The 0-valued level set of $\psi^k(x,W)$ is the classification boundary between label $k$ and all others. Summing the total robustness over all different $\psi^k$'s give us the multilabel topological penalty. 
We omit the derivation of the gradients due to space constraint.
The computation is similar to binary-labeled setting, except that at each iteration, we need to compute the persistence pairs for all the $K$ functions.

%
\section{Experiments}
\label{sec:exp}
We test our method (TopoReg) on multiple synthetic datasets and real world datasets. 
The weight of the topological penalty, $\lambda$ and the Gaussian kernel width $\sigma$ are turned via cross-validation.
To compute topological information requires discretization of the domain. For 2D data, we normalize the data to fit the unit square $[0,1]\times[0,1]$, and discretize the square into a grid with 300$\times$ 300 vertices. For high-dimensional data, we use the KNN graph with $k = 3$. 

\begin{table*}[h!]
    \caption{The mean and standard deviation error rate of different methods.} 
    \label{tab:exp}
    \centering

        \begin{tabular}{l|c |c | c|c |c | c |c }
            \toprule
            \hline

            \multicolumn{8}{c}{Synthetic}  \\
            \hline

            & KNN & LG & SVM & EE &  DGR & KLR  &   TopoReg   \\
            \hline

            \hline


            Blob-2 (500,5) & 7.61 & 8.20 & 7.61  & 8.41 & 7.41  & 7.80  &    \bf{7.20}   \\
            \hline
            Moons (500,2) & 20.62 & 20.00 & 19.80  & 19.00  & 19.01   & 18.83  &   \bf{18.63}  \\
            \hline
            Moons (1000,2,Noise 0\%) & 19.30 & 19.59  & 19.89 & 17.90 & 19.20 & 17.80 & \bf{17.60} \\
            \hline
            Moons (1000,2,Noise 5\%) & 21.60 & 19.29 & 19.59 & 22.00  &  22.30 & \bf{19.00} & \bf{19.00} \\
            \hline
            Moons (1000,2,Noise 10\%) & 21.10 & \bf{19.19} &  19.89 & 24.40 & 26.30 & 20.00 & {19.70} \\
            \hline
            Moons (1000,2,Noise 20\%) & 23.00 & 19.79 & \bf{19.40} & 30.60 & 30.20 & 19.50 & \bf{19.40} \\
            \hline
            AVERAGE  & 18.87 & 17.68 & 17.70 & 20.39 & 20.74 & 21.63 & \bf{16.92}\\
            \hline
            \hline
            \multicolumn{8}{c}{UCI}  \\
            \hline
            & KNN & LG &  SVM & EE &  DGR & KLR  &   TopoReg   \\
            \hline
            SPECT (267,22) & 17.57 & 17.20 & 18.68  & \bf{16.38}   & 23.92  & 18.31  & {17.54} \\
            \hline
            Congress (435,16) & 5.04 & 4.13 &  4.59  & 4.59 & 4.80  & \bf{4.12}  & 4.58  \\
            \hline

            Molec. (106,57) & 24.54  &   19.10 & 19.79  & 17.25  & 16.32  & 19.10  & \bf{12.62}  \\
            \hline

             Cancer (286,9)  & 29.36  &  28.65  & 28.64  & 28.68  & 31.42  & 29.00  &\bf{28.31}  \\
            \hline


            Vertebral (310,6) & 15.47  &  15.46 & 23.23 & 17.15 & 13.56  &  12.56  & \bf{12.24}  \\
            \hline

            Energy (768,8) & 0.78  &  0.65 & 0.65  & 0.91   & 0.78  & \bf{0.52} & \bf{0.52}  \\
            \hline
            AVERAGE & 15.46 & 14.20  & 15.93 & 14.16 & 15.13 & 13.94 & \bf{11.80}\\

            \hline
            \hline
            \multicolumn{8}{c}{Biomedicine}  \\
            \hline
            & KNN & LG  &  SVM & EE &  DGR & KLR  &   TopoReg   \\
            \hline
            KIRC (243,166) & 30.12  &  28.87  & 32.56 & 31.38   & 35.50  & 31.38  &  \bf{26.81}  \\
            \hline
            fMRI (1092,19) & 46.70  &  74.91  & 74.08 & 82.51 &  \bf{31.32} & 34.07 & 33.24 \\




            \hline
            \bottomrule
        \end{tabular}
\end{table*}

\begin{figure}[btp!]
\vspace{-.1in}
  \centering
    \includegraphics[width=.25\textwidth]{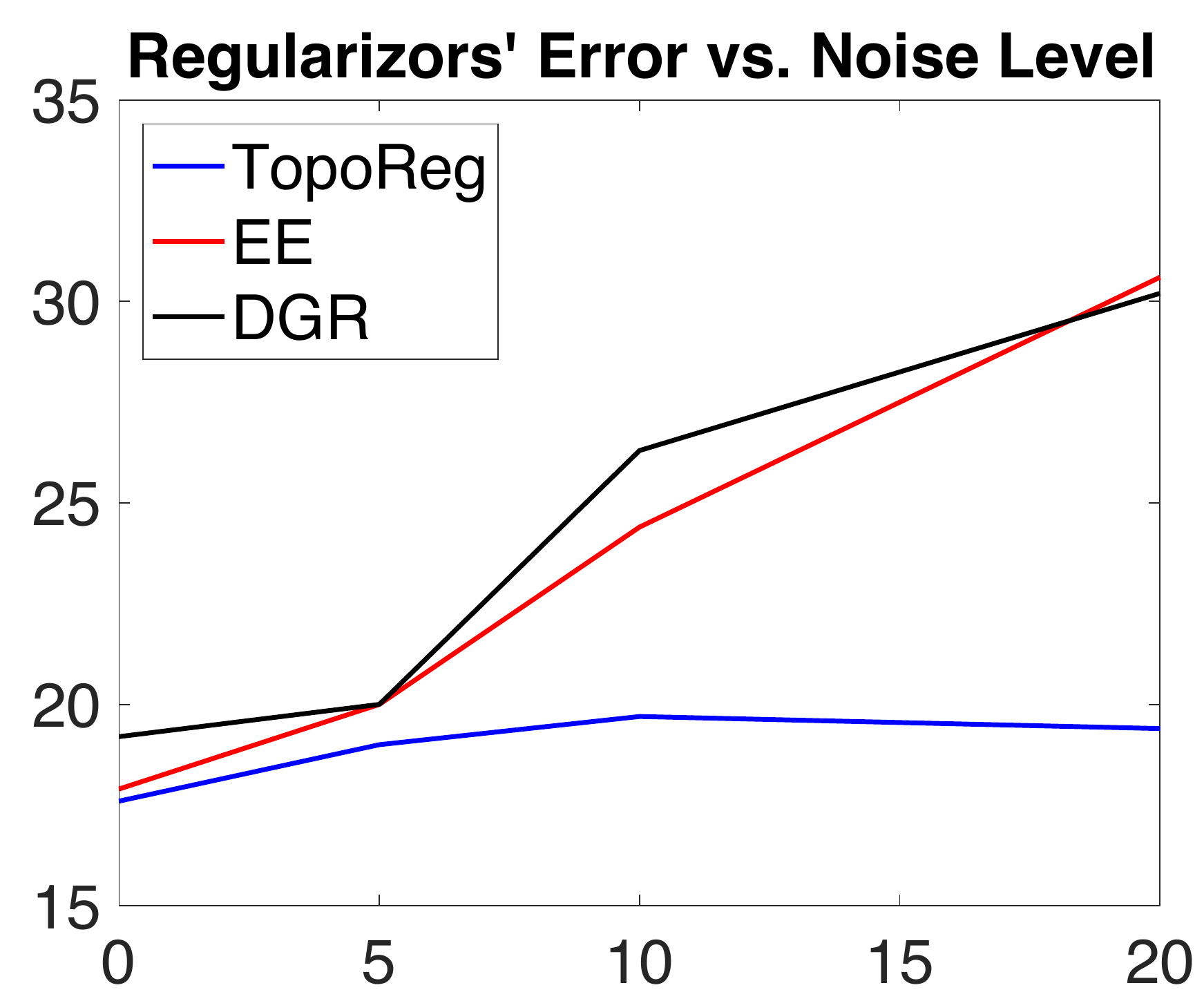} 
\vspace{-.1in}
  \caption{Comparison of different regularizers with different noise level. TopoReg is more robust to the noise compared with other geometric regularizers.}
\label{fig:comp-noise}
\vspace{-.1in}
\end{figure}

\myparagraph{Baselines.}
We compare our method with several baselines: k-nearest-neighbor classifier (KNN), logistic regression (LG), Support Vector Machine (SVM), and Kernel Logistic Regression (KLR) with functional norms ($L_1$ and $L_2$) as regularizers. 
We also compare with two state-of-the-art methods based on geometric-regularizers: the Euler’s Elastica classifier (EE) \citep{Lin2015} and the Classifier with Differential Geometric Regularization (DGR) \citep{bai2016differential}.  
All relevant hyperparameters are tuned using cross-validation.

For every dataset and each method, we randomly divide the datasets into 6 folds. Then we use each of the 6 folds as testing set, while doing a 5-fold cross validation on the rest 5 folds data to find the best hyperparameters. Once the best hyperparameters are found, we train on the entire 5 folds data and test on the testing set. 

\myparagraph{Data.}
In order to thoroughly evaluate the behavior of our model, especially in large noise regime, we created synthetic data with various noise levels.
Beside feature space noise, we also inject different levels of label noise, \eg, randomly perturb labels of 0\%, 5\%, 10\% and 20\% of the training data.
We also evaluate our method on real world data.
We use several UCI datasets with various sizes and dimensions to test our method \citep{Lichman2013}. In addition, we use two biomedical datasets. The first is the kidney renal clear cell carcinoma cancer (KIRC) dataset \citep{yuan2014assessing} extracted from the Cancer Genome Atlas project (TCGA) \citep{TCGA}. The features of the dataset are the protein expression measured on the MD Anderson Reverse Phase Protein Array Core platform (RPPA). 
The second dataset is a task-evoked functional MRI images, which has 19 dimensions (corresponding to activities at 19 brain ROIs) and 6 labels (corresponding to 6 different tasks) \citep{ni2018region}.

The results are reported in Table~\ref{tab:exp}. We also report the average performance over each category (AVERAGE). The two numbers next to each dataset name are the data size $N$ and the dimension $D$, respectively. The average running time over all the datasets for our method is 2.08 seconds. 

\myparagraph{Discussions.}
Our method generally outperforms existing methods on datasets in Table \ref{tab:exp}. More importantly, we note that our method also provides consistently best or close to best performance among all approaches tested. (For example, while EE performs well on some datasets, its performance can be significantly worse than the best for some other datasets.) 

On synthetic data, we found that TopoReg has a bigger advantage on relatively noisy data. 
This is expected. Our method provides a novel way to topologically simplify the global structure of the model, without having to sacrifice too much of the flexibility of the model. Meanwhile, to cope with large noise, other baseline methods have to enforce an overly strong global regularization in a structure agnostic manner.
We also observe that TopoReg performs relatively stable when label noise is large, while the other geometric regularizers are much more sensitive to label noise. See Figure \ref{fig:comp-noise} for a comparison. We suspect this is because the other geometric regularizers are more sensitive to the initialization and tend to stuck in bad local optima.




%
\bibliographystyle{apalike}
\bibliography{toporeg,topovis,deeplearning,nipspaper,tda,kernel,cite}

\appendix
\section{Background: Persistent Homology}
\label{app:pers}
Persistent homology~\citep{ELZ02,zomorodian2005computing,CS10,CSM09} is a fundamental recent development in the field of computational topology, underlying many topological data analysis methods.
Below, we provide an an intuitive description to help explain its role in measuring the robustness of topological features in the zero-th level set (the separation boundary) of classifier function $f$. 

\begin{figure}[hbtp]
\begin{tabular}{cc}
\includegraphics[height=3cm]{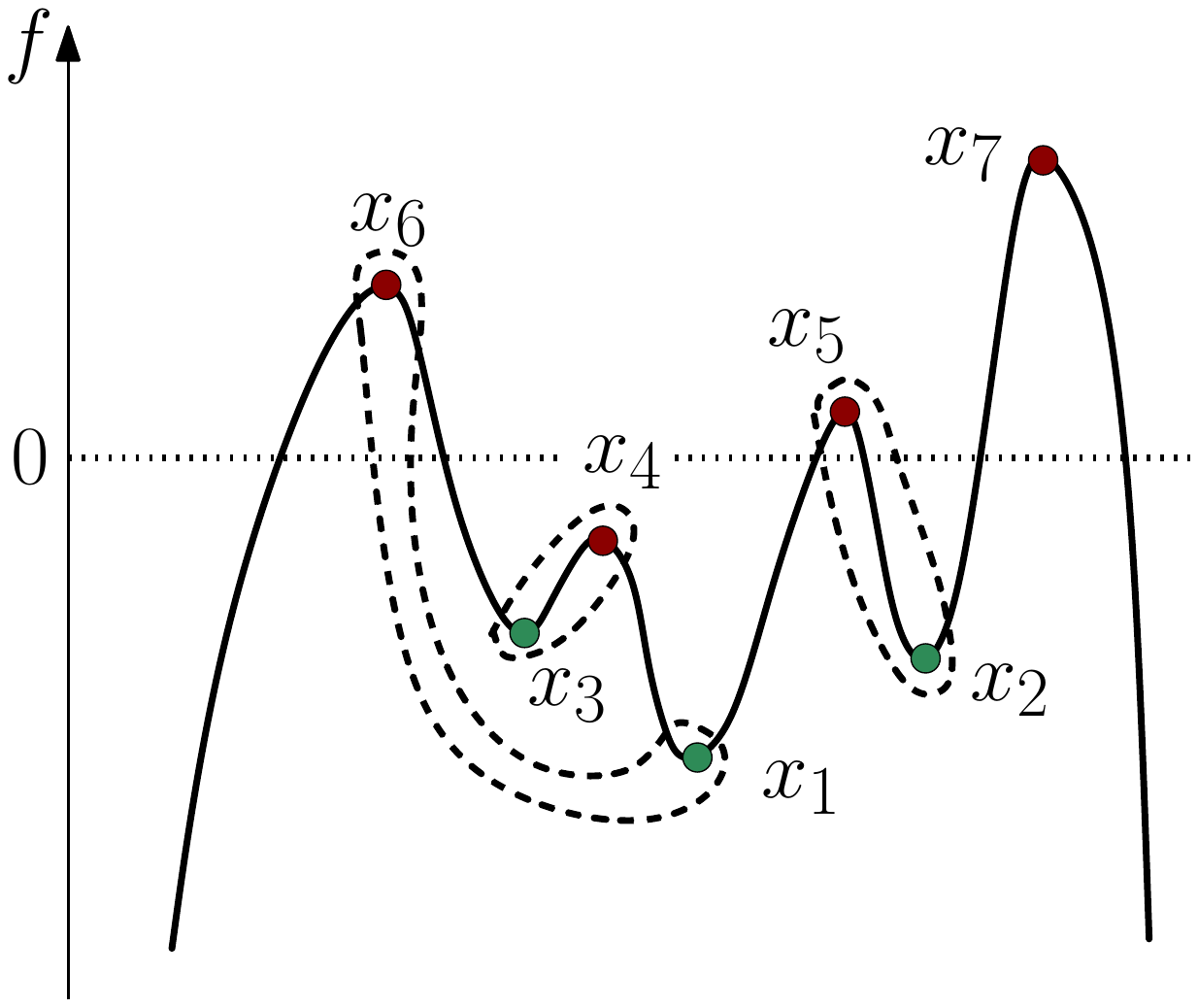} &
\hspace{-.2in}
\includegraphics[height=2.5cm]{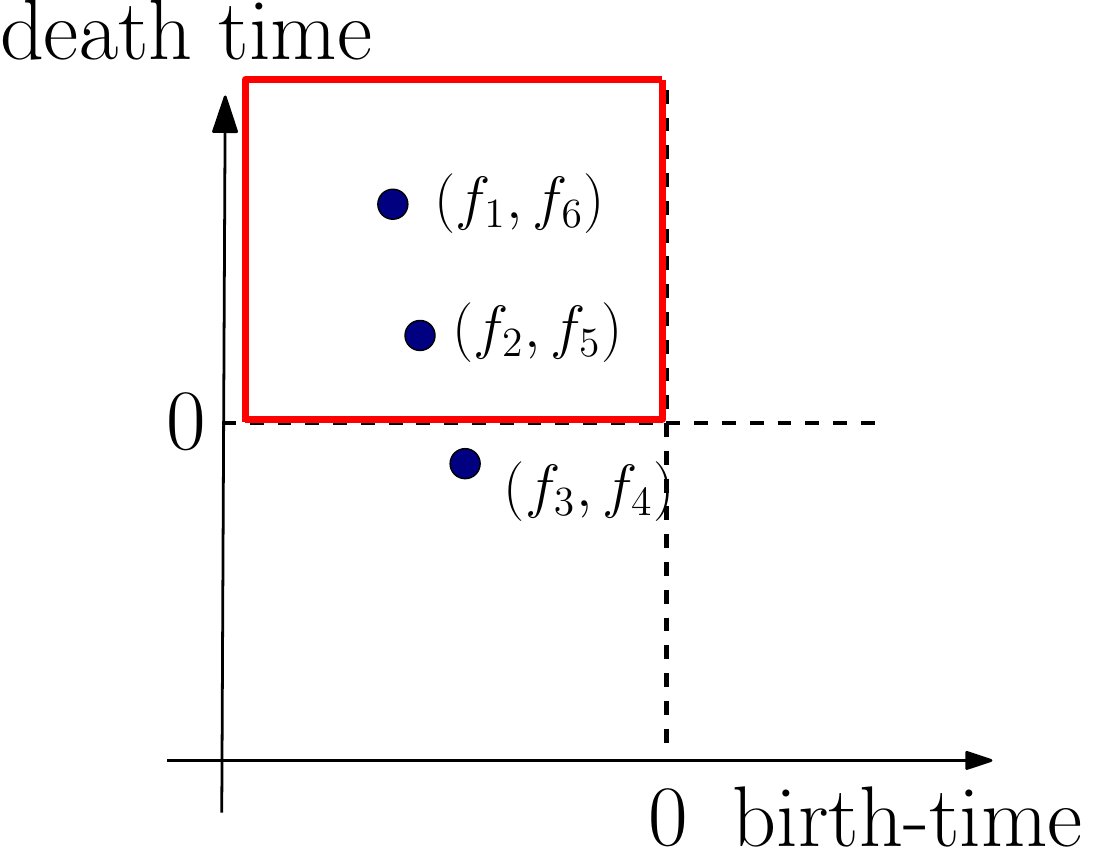}\\
(a) & (b)
\end{tabular}
\caption{(a) A function $f: \R \to \R$. Its persistence pairings (of critical points) are marked by the dotted curves: $\Pi_f = \{(x_1, x_6), (x_2, x_5), (x_3, x_4), \ldots \}$. 
    The corresponding persistence diagram is shown in (b), with $\dgm(f) = \{ (f_1, f_6), (f_2, f_5), (f_3, f_4), \ldots \}$, where $f_i  = f(x_i)$ for each $i\in [1, 6]$. 
For example, as we sweep pass minimum $x_3$, a new component is created in the sub-level set. This component is merged to an older component (created at $x_1$) when we sweep past critical point (maximum) $x_4$. This gives rise to a persistence pairing $(x_3, x_4)$ corresponding to the point $(f_3, f_4)$ in the persistence diagram. 
\label{fig:1Dper}}
\end{figure}

Suppose we are given a space $Y$ and a continuous function $f: Y \to \R$ defined on it. 
To characterize $f$ and $Y$, imagine we now sweep the domain $Y$ in increasing $f$ values. 
This gives rise to the following growing sequence of sublevel sets: 
$$Y_{\le t_1} \subseteq Y_{\le t_2} \subseteq \cdots \subseteq Y_{\le t_m}, 
\text{with}~~ t_1 < t_2 < \cdots t_m, $$
where $Y_{\le t}:= \{ x \in Y \mid f(x) \le t \}$ is the \emph{sublevel set of $f$ at $t$}. 
We call it the \emph{sublevel set filtration of $Y$ \wrt $f$}, which intuitively inspects $Y$ from the point of view of function $f$. 
During the sweeping process, sometimes, new topological features (homology classes), say, a new component or a handle, will be created. Sometimes an existing one will be killed, say a component either disappear or merged into another one, or a void is filled. 
It turns out that these changes will only happen when we sweep through a critical points of the function $f$. 
The persistent homology tracks these topological changes, and pair up critical points into a collection of \emph{persistence pairings} $\Pi_f = \{ (b, d) \}$. Each pair $(b, d)$ are the critical points where certain topological feature is created and killed. Their function values $f(b)$ and $f(d)$ are referred to as the \emph{birth time} and \emph{death time} of this feature.
The corresponding collection of pairs of (birth-time, death-time) is called the \emph{persistence diagram}, formally, \mbox{$\dgm(f) = \{ (f(b), f(d)) \mid (b, d) \in \Pi\}$}. 
For each persistent pairing $(b, d)$, its \emph{persistence} is defined to be $|f(d) - f(b)|$, which measures the life-time (and thus importance) of the corresponding topological feature \wrt $f$. 
A simple 1D example is given in Figure \ref{fig:1Dper}. 

The above description is the tranditional persistence (induced by the sub-level set filtratoin of $f$) introduced in \cite{ELZ02}, which we refer to as \emph{ordinary persistence} in what follows. 
To capture all the topological features in the \emph{level sets} (instead of sublevel sets) of all different threshold values, we use an extension of the aforementioned sublevel set persistence, called the \emph{level set zigzag persistence} \citep{CSM09}. Intuitively, we sweep the domain $Y$ in increasing function values and now track topological features of the level sets, instead of the sublevel sets. 
The resulting set of persistence pairings $\Pi_Z(f)$ and persistence diagram $\dgm_Z(f)$ have analogous meanings: each pair of critical points $(b, d)\in \Pi_Z(f)$ corresponds to the creation and killing of some homological feature (e.g, connected components if we look at 0-th dimensional homological features) in the level set, and the corresponding pair $(f(b), f(d))\in \dgm(f)$ are the birth / death times of this feature.

\myparagraph{Sketch of proof for Theorem 2.1.} 
Now for a classifier function $f: \calX \to \R$, given the level set zigzag persistence diagram $\dgm_Z(f)$ and its corresponding set of persitsence pairings $\Pi_Z(f)$ w.r.t. $f$, we collect $\Pi := \{ (p, q) \in \Pi_Z(f) \mid f(p) \le 0, f(q) \ge 0\}$. 
Intuitively, each $(p, q)\in \Pi$ corresponds to a 0-D homological feature (connected component) that first appeared in a level set $f^{-1}(a)$ with $a = f(p) \le 0$ below the zero level set $f^{-1}(0)$, and it persists through the zero level set and dies only in the level set w.r.t. value $f(q) \ge 0$. 
Thus each $(p, q)$ corresponds to a distinct connected component $c$ in the zero level set $S_f = f^{-1}(0)$. 
Hence intuitively, this set of persistence pairings $\Pi$ maps to the set of connected components in the separation boundary $S_f$ \emph{bijectively} as claimed in Theorem 2.1 \footnote{We note that one can argue this more formally by considering the 0th-dimensional levelset zigzag persistence module, and its decomposition into the interval modules \cite{CS10}. The rank of the 0th homology group of a specific levelset, say $H_0(f^{-1}(a))$, equals to the number of interval modules whose span covers $a$. }. 
%
%
To remove this component from the zero level set $S_f$, we either need to move down $q$ from $f(q)$ to $0$, or move up $p$ from $f(p)$ to $0$. The robustness of this component $c$ is thus $\rho(c) = \min \{ | f(p)|, |f(q)| \}$. 

In general, the level set zigzag persistence takes $O(n^3)$ time to compute \cite{CSM09}, where $n$ is the total complexity of the discretized representation of the domain $\calX$. 
However, first, we only need the 0th dimensional levelset zigzag persistence. Furthermore, our domain $\calX$ is a hypercube (thus simply connected). 
Using Theorem 2 of \cite{bendich2013homology}, and combined with the {\sf EP Symmetry Corollary} of \cite{CSM09}, one can then show the following: 

Let $\dgm(f)$ and $\dgm(-f)$ denote the ordinary 0-dimensional persistence diagrams w.r.t. the sublevel set filtrations of $f$ and of $-f$, respectively. Let $\Pi(f)$ and $\Pi(-f)$ denote their corresponding set of persistence pairings. 
Set $\widehat{\Pi}_f:= \{ (p, q) \in \Pi(f) \mid f(p) \le 0, f(q) \ge 0\}$ and $\widehat{\Pi}_{-f} :=\{ (p, q) \in \Pi(-f) \mid -f(p) \le 0, -f(q) \ge 0\}$. 
(For example, in Figure \ref{fig:1Dper}(b), points in the red box correspond to $\widehat{\Pi}_f$.
With this correspondence, for each 0D topological feature (connected component) $c$ from $S$, let $(b_c, d_c)$ be its pairing of critical points. The corresponding $(f(b_c),f(d_c))$ belongs to $\widehat{\dgm}$.
Given a persistence pair $(p, q)$, we say that the range of this pair covers $0$ if $f(p) \le 0 \le f(q)$. 
Then by Theorem 2 of \cite{bendich2013homology}, and combined with the {\sf EP Symmetry Corollary} of \cite{CSM09}, we have that 
$$\Pi = \widehat{\Pi}_f \cup \widehat{\Pi}_{-f} \cup \widehat{\Pi}_E, $$ 
where $\widehat{\Pi}_E$ consists certain persistence pairs (whose range covers $0$) from the 0-th and 1-st dimensional \emph{extended subdigrams} induced by the extended persistent homology \cite{CEH09}. 
However, since $\calX$ is simply connected, $H_1(\calX)$ is trivial. Hence there is no point in the 1-st extended subdigram. As $\calX$ is connected, there is only one point $\{(v_1, v_n)\}$ in the 0-th extended subdigram, where $v_1$ and $v_n$ are the global minimum and global maximum of the function $f$, respectively. 
It then follows that 
\begin{align}\label{eqn:perpairs}
\Pi = \widehat{\Pi}_f \cup \widehat{\Pi}_{-f} \cup \{ (v_1, v_n) \}. 
\end{align}
Hence one can compute $\Pi$ by computing the 0-th ordinary persistence homology induced by the sublevel set filtration of $f$, and of $-f$, respectively. 
This finishes the proof of Theorem 2.1. 


\begin{remark}
Finally, we can naturally extend the above definition by considering persistent pairs and the diagram corresponding to the birth and death of high dimensional topological features, \eg, handles, voids, \etc.  
\label{rem:complexity}
\end{remark}

%
%
%
%
%
%
%
%
\end{document}